# Marginalization in Composed Probabilistic Models

Radim Jiroušek
Laboratory for Intelligent Systems
University of Economics, Prague
and
Institute of Information Theory and Automation
Academy of Sciences of the Czech Republic
Email: radim@vse.cz

## Abstract

Composition of low-dimensional distributions, whose foundations were laid in the paper published in the Proceedings of UAI'97 (Jiroušek 1997), appeared to be an alternative apparatus to describe multidimensional probabilistic models. In contrast to Graphical Markov Models, which define multidimensional distributions in a declarative way, this approach is rather procedural. Ordering of low-dimensional distributions into a proper sequence fully defines the respective computational procedure; therefore, a study of different types of generating sequences is one of the central problems in this field. Thus, it appears that an important role is played by special sequences that are called *perfect*. Their main characterization theorems are presented in this paper. However, the main result of this paper is a solution to the problem of marginalization for general sequences. The main theorem describes a way to obtain a generating sequence that defines the model corresponding to the marginal of the distribution defined by an arbitrary generating sequence. From this theorem the reader can see to what extent these computations are *local*; i.e., the sequence consists of marginal distributions whose computation must be made by summing up over the values of the variable eliminated (the paper deals with a finite model).

## 1 INTRODUCTION

Bayesian networks, perhaps the most famous model for representation of multidimensional probability distributions, belong to a wider class of models that are most often called Graphical Markov models. All of these models are proposed to represent distributions of high dimensionality (hundreds or even thousands dimensions), which cannot generally be handled because of the exponential growth of the number of necessary parameters. What is common to all these models is the fact that they can represent distributions with special dependence structures (namely, this feature makes it possible to define the distribution with the aid of a moderate number of parameters), and that these structures are described by graphs. The approach presented herein keeps the former property, abandoning the latter.

Instead of representing the dependence structure of a modeled distribution, the presented approach describes the computational process that defines the multidimensional distribution. Naturally, one can see the dependence structure from this process. But it is not the main goal of this apparatus.

Our approach is based on the operators of composition, used for construction of the distribution of variables $(X_i)_{i \in K_1 \cup K_2}$ from two low-dimensional distributions, $P_1(X_i)_{i \in K_1}$ and $P_2(X_i)_{i \in K_2}$. These operators, as well as their basic properties, are introduced in the next section. The third section describes the main idea, namely, generating the multidimensional distributions by iterative application of these operators; and the most important generating sequences, called perfect sequences, are characterized. The fourth section is devoted to the main focus of the paper: marginalization of multidimensional distributions defined by generating sequences.

## 2 NOTATION AND BASIC PROPERTIES

In this paper, we will deal with probability distributions $P((X_i)_{i \in K})$, where both the index set $K$ and all sets of values of variables $X_i$ are assumed to be finite. Different distributions can be (and usually are) defined for different sets of variables. To simplify the notation as much as possible, distributions $P_1, P_2, \ldots, P_n$ will always be defined for variables whose indices lie in sets $K_1, K_2, \ldots, K_n$, respectively. In other words, we are going to consider distributions

$$P_1((X_i)_{i \in K_1}), P_2((X_i)_{i \in K_2}), \ldots, P_n((X_i)_{i \in K_n}).$$

Having a distribution $P$ (i.e. $P((X_i)_{i \in K})$), we will often consider its marginal distributions. For $L \subset K$, the marginal distribution $P((X_i)_{i \in L})$ will be denoted by using set $L$ as an upper index in round parentheses: $P^{(L)}$. Considering a general $L$ (i.e., $L$ is not necessarily a part of $K$), the symbol $P^{(L)}$ will denote the marginal distribution $P((X_i)_{i \in K \cap L})$. Two distributions $P_1$ and $P_2$ are called *consistent* if

$$P_1^{(K_1 \cap K_2)} = P_2^{(K_1 \cap K_2)}.$$

The main theorem of this paper deals with marginalizing one variable out, i.e., it describes a form of the marginal distribution $P^{(K \setminus \{\ell\})}$ for $\ell \in K$. To simplify the notation, for this type of a marginal distribution we shall use the symbol $P^{[\ell]}$.

The most important concept of this contribution is that of the *composition* operators. To make it clear from the very beginning, let us stress that it is nothing else but a generalization of the idea of computing a three-dimensional distribution from two two-dimensional ones by introducing the conditional independence of variables $X_1$ and $X_3$ given $X_2$:

$$P_1(X_1, X_2) \triangleright P_2(X_2, X_3) = \frac{P_1(X_1, X_2) P_2(X_2, X_3)}{P_2(X_2)}$$
$$= P_1(X_1, X_2) P_2(X_3 | X_2).$$

Consider two probability distributions $P_1((X_i)_{i \in K_1})$ and $P_2((X_i)_{i \in K_2})$. A *right composition* of probability distributions $P_1$ and $P_2$ is defined by the formula

$$P_1 \triangleright P_2 = \begin{cases} \frac{P_1 P_2}{P_2^{(K_1 \cap K_2)}} & \text{if } P_1^{(K_1 \cap K_2)} \ll P_2^{(K_1 \cap K_2)}, \\ \text{undefined} & \text{otherwise,} \end{cases}$$

where $\left( P_1^{(K_1 \cap K_2)} \ll P_2^{(K_1 \cap K_2)} \right)$ means that

$$\left( P_2((X_i = x_i)_{i \in (K_1 \cap K_2)}) \right) = 0$$
$$\implies P_1((X_i = x_i)_{i \in (K_1 \cap K_2)}) = 0).$$

Taking $\frac{0 \cdot 0}{0} = 0$, the operation of composition, if defined, results in a probability distribution

$$(P_1 \triangleright P_2)((X_i)_{i \in K_1 \cup K_2})$$

and its marginal distribution $(P_1 \triangleright P_2)((X_i)_{i \in K_1})$ (or, using the symbol more often utilized in the sequel,

$(P_1 \triangleright P_2)^{(K_1)}$) equals $P_1$. If $K_1 \cap K_2 = \emptyset$, then $P_1 \triangleright P_2$ degenerates to a simple product of $P_1$ and $P_2$.

Analogously, we can also introduce the operator of *left composition*:

$$P_1 \triangleleft P_2 = \begin{cases} \frac{P_1 P_2}{P_1^{(K_1 \cap K_2)}} & \text{if } P_2^{(K_1 \cap K_2)} \ll P_1^{(K_1 \cap K_2)}, \\ \text{undefined} & \text{otherwise,} \end{cases}$$

These operators, when applied iteratively, construct multidimensional distributions from a set of low-dimensional ones. In this paper we will primarily concentrate on sequences connected by the operator of right composition:

$$P_1 \triangleright P_2 \triangleright \ldots \triangleright P_n.$$

This formula, if it is defined, determines the distribution of variables $(X_i)_{i \in K_1 \cup K_2 \cup \ldots \cup K_n}$. Regarding the fact (see Jiroušek 1997) that the operator $\triangleright$ is neither commutative nor associative, we must stress just this once that we always apply the operators from left to right:

$$P_1 \triangleright P_2 \triangleright \ldots \triangleright P_n = (\ldots ((P_1 \triangleright P_2) \triangleright P_3) \triangleright \ldots \triangleright P_n).$$

As can already be seen from the above formulations, when speaking about properties of generating sequences we often have to distinguish between the situations in which the respective formulae are or are not defined. Describing singular situations with undefined formulae is, from the point of view of this paper, quite uninteresting. To avoid the necessity of repeating technical exercises on each occasion, let us assume that all the formulae are well defined. It can, for example, be guaranteed by an assumption that all the distributions $P_k$ are positive.

Now, let us introduce a couple of assertions that will be necessary in the next sections. The first two lemmata were proven in (Jiroušek 1997).

**Lemma 1** $P_1$ and $P_2$ are consistent iff

$$P_1 \triangleright P_2 = P_1 \triangleleft P_2. \qquad \square$$

**Lemma 2** If $K_1 \supseteq (K_2 \cap K_3)$ then

$$P_1 \triangleright P_2 \triangleright P_3 = P_1 \triangleright P_3 \triangleright P_2. \qquad \square$$

**Lemma 3** Let $L$ be such that $K_1 \cap K_2 \subseteq L \subseteq K_2$; then

$$P_1 \triangleright P_2 = P_1 \triangleright P_2^{(L)} \triangleright P_2.$$

*Proof.* The assertion follows immediately from the definition of the composition operator $\triangleright$:

$$P_1 \triangleright P_2^{(L)} \triangleright P_2 = \frac{P_1 P_2^{(L)}}{P_2^{(L \cap K_1)}} \frac{P_2}{P_2^{(K_2 \cap (K_1 \cup L))}}$$

$$= \frac{P_1 P_2^{(L)}}{P_2^{(K_1 \cap K_2)}} \frac{P_2}{P_2^{((K_2 \cap K_1) \cup L)}}$$

$$= \frac{P_1 P_2^{(L)}}{P_2^{(K_1 \cap K_2)}} \frac{P_2}{P_2^{(L)}}$$

$$= \frac{P_1 P_2}{P_2^{(K_1 \cap K_2)}} = P_1 \triangleright P_2.$$

$\square$

The following theorem is of ultimate importance for the assertions from Section 4. There are two additional reasons for our presentation of its proof, despite its having already been proven in (Jiroušek 1997). First, the proof presented here is more transparent than the original one, and second, there is a certain license in defining the operator $\odot_K$, which appears in the assertion. This arbitrariness, which will be discussed in more detail below, can be seen from the proof.

**Theorem 1**

$$P_1 \triangleright P_2 \triangleright P_3 = P_1 \triangleright (P_2 \odot_{K_1} P_3) = P_2 \odot_{K_1} P_3 \triangleleft P_1,$$

*where*

$$P_2 \odot_{K_1} P_3 = (P_3^{((K_1 \setminus K_2) \cap K_3)} P_2) \triangleright P_3.$$

*Proof.*

$$P_2 \odot_{K_1} P_3 = (P_3^{((K_1 \setminus K_2) \cap K_3)} P_2) \triangleright P_3$$

$$= \frac{P_3^{((K_1 \setminus K_2) \cap K_3)} P_2 P_3}{P_3^{((K_1 \cup K_2) \cap K_3)}}.$$

Therefore

$$P_1 \triangleright (P_2 \odot_{K_1} P_3) = \frac{P_1 \frac{P_3^{((K_1 \setminus K_2) \cap K_3)} P_2 P_3}{P_3^{((K_1 \cup K_2) \cap K_3)}}}{\left(\frac{P_3^{((K_1 \setminus K_2) \cap K_3)} P_2 P_3}{P_3^{((K_1 \cup K_2) \cap K_3)}}\right)^{((K_2 \cup K_3) \cap K_1)}}$$

$$= \frac{P_3^{((K_1 \setminus K_2) \cap K_3)} \frac{P_1 P_2 P_3}{P_3^{((K_1 \cup K_2) \cap K_3)}}}{P_3^{((K_1 \setminus K_2) \cap K_3)} \left(\frac{P_2 P_3}{P_3^{((K_1 \cup K_2) \cap K_3)}}\right)^{((K_2 \cup K_3) \cap K_1)}}$$

$$= \frac{\frac{P_1 P_2 P_3}{P_3^{((K_1 \cup K_2) \cap K_3)}}}{\left(\frac{P_2 P_3}{P_3^{((K_1 \cup K_2) \cap K_3)}}\right)^{((K_2 \cup K_3) \cap K_1)}},$$

where the second modification is feasible because

$$(K_1 \setminus K_2) \cap K_3 \subseteq (K_2 \cup K_3) \cap K_1.$$

Notice here that the last modification is just an elimination of the auxiliary distribution $P_3^{((K_1 \setminus K_2) \cap K_3)}$ introduced in the definition of the operator $\odot_{K_1}$.

Let us focus our attention on the denominator of the last fraction. It is a marginal of a product of $P_2$ with a conditional distribution $P_3((X_i)_{i \in K_3 \setminus (K_1 \cup K_2)} | (X_i)_{i \in K_3 \cap (K_1 \cup K_2)})$. When computing this marginal, we have to sum up over all combinations of values of variables $(X_i)_{i \in (K_2 \cup K_3) \setminus K_1}$. In the following computations we will separate these variables into two groups: $(X_i)_{i \in K_2 \setminus K_1}$ and $(X_i)_{i \in K_3 \setminus (K_1 \cup K_2)}$. Let $\mathbf{X}_{K_2 \setminus K_1}$ and $\mathbf{X}_{K_3 \setminus (K_1 \cup K_2)}$ be the sets whose elements are all combinations of values of variables $(X_i)_{i \in K_2 \setminus K_1}$ and $(X_i)_{i \in K_3 \setminus (K_1 \cup K_2)}$, respectively. $x \in \mathbf{X}_{K_2 \setminus K_1}$ is thus a vector of values of variables $(X_i)_{i \in K_2 \setminus K_1}$, with $x_i$ denoting the coordinate which corresponds to the value of variable $X_i$. Analogously, $y \in \mathbf{X}_{K_3 \setminus (K_1 \cup K_2)}$ is a vector of values of variables $(X_i)_{i \in K_3 \setminus (K_1 \cup K_2)}$ and $y_i$ again denotes the corresponding value of variable $X_i$. Using this notation, we can compute:

$$(P_2((X_i)_{i \in K_2}) P_3((X_i)_{i \in K_3 \setminus (K_1 \cup K_2)} | (X_i)_{i \in K_3 \cap (K_1 \cup K_2)}))^{((K_2 \cup K_3) \cap K_1)}$$

$$= \sum_{x \in \mathbf{X}_{K_2 \setminus K_1}} \sum_{y \in \mathbf{X}_{K_3 \setminus (K_1 \cup K_2)}} P_2((X_i)_{i \in K_2 \cap K_1}, (X_i = x_i)_{i \in K_2 \setminus K_1}) P_3((X_i = y_i)_{i \in K_3 \setminus (K_1 \cup K_2)} | (X_i = x_i)_{i \in (K_3 \cap K_2) \setminus K_1}, (X_i)_{i \in K_3 \cap K_1})$$

$$= P_2((X_i)_{i \in K_2 \cap K_1}) \sum_{x \in \mathbf{X}_{K_2 \setminus K_1}} P_2((X_i = x_i)_{i \in K_2 \setminus K_1} | (X_i)_{i \in K_2 \cap K_1}) \sum_{y \in \mathbf{X}_{K_3 \setminus (K_1 \cup K_2)}} P_3((X_i = y_i)_{i \in K_3 \setminus (K_1 \cup K_2)} | (X_i = x_i)_{i \in (K_3 \cap K_2) \setminus K_1}, (X_i)_{i \in K_3 \cap K_1})$$

$$= P_2((X_i)_{i \in K_2 \cap K_1})$$

Substituting this result back into the denominator of the fraction, we get

$$P_1 \triangleright (P_2 \odot_{K_1} P_3) = \frac{\frac{P_1 P_2 P_3}{P_3^{((K_1 \cup K_2) \cap K_3)}}}{P_2^{(K_2 \cap K_1)}}$$

$$= \frac{P_1 P_2 P_3}{P_2^{(K_2 \cap K_1)} P_3^{((K_1 \cup K_2) \cap K_3)}}$$

$$= P_1 \triangleright P_2 \triangleright P_3.$$

which completes the proof. $\square$

As we mentioned previously, we could define the $\otimes_K$ operator with the aid of an (almost) arbitrary distribution $R$

$$P_2 \otimes_{K_1} P_3 = (R^{((K_1 \setminus K_2) \cap K_3)} P_2) \triangleright P_3.$$

For example, an arbitrary positive distribution which is defined for the respective variables will serve well. For the sake of simplicity, it seems reasonable to consider a uniform distribution. The specific purpose of this distribution is simply to introduce the necessary conditional independence that would otherwise be omitted. To illustrate the point, let us consider the following trivial example:

$$P_1(X_1) \triangleright P_2(X_2) \triangleright P_3(X_1, X_2) = P_1(X_1) P_2(X_2).$$

If we used the operator $\triangleright$ instead of $\otimes_{K_1}$, we would get

$$\begin{aligned} &P_1(X_1) \triangleright (P_2(X_2) \triangleright P_3(X_1, X_2)) \\ &= \frac{P_1(X_1)(P_2(X_2) P_3(X_1|X_2))}{\sum_{x \in \mathbf{X}_1} P_2(X_2) P_3(X_1 = x|X_2)}, \end{aligned}$$

which evidently differs from $P_1(X_1) P_2(X_2)$ because $P_1 \triangleright (P_2 \triangleright P_3)$ inherits the dependence of variables $X_1$ and $X_2$ from $P_3$. Nevertheless, considering

$$\begin{aligned} &P_1(X_1) \triangleright (P_2(X_2) \otimes_{\{1\}} P_3(X_1, X_2)) \\ &= P_1(X_1) \triangleright (P_3(X_1) P_2(X_2) \triangleright P_3(X_1, X_2)) \\ &= P_1(X_1) \triangleright P_3(X_1) P_2(X_2) \\ &= P_1(X_1) P_2(X_2) \end{aligned}$$

gives the desired result.

## 3 GENERATING SEQUENCES

Using operators of composition, we can construct multidimensional distributions from a system of low-dimensional ones. As a rule, we consider constructions that apply one of the two introduced operators iteratively. This means we consider either distributions

$$P_1 \triangleright P_2 \triangleright \ldots \triangleright P_n,$$

or

$$P_1 \triangleleft P_2 \triangleleft \ldots \triangleleft P_n.$$

Since these formulae generally define different distributions, it is reasonable to study both of them. However, though it is perhaps not evident at first sight, these two expressions substantially differ from each other, namely, from the computational point of view. Consider an index $k \in \{1, 2, \ldots, n-1\}$ which is close to $n-1$. Application of the $k$-th operator means the computation of either

$$(P_1 \triangleright \ldots \triangleright P_k) \triangleright P_{k+1} = \frac{(P_1 \triangleright \ldots \triangleright P_k) P_{k+1}}{P_{k+1}^{(K_{k+1} \cap (K_1 \cup \ldots \cup K_k))}}$$

for the application of the operator $\triangleright$; or

$$\begin{aligned} &(P_1 \triangleleft \ldots \triangleleft P_k) \triangleleft P_{k+1} \\ &= \frac{(P_1 \triangleleft \ldots \triangleleft P_k) P_{k+1}}{(P_1 \triangleleft \ldots \triangleleft P_k)^{(K_{k+1} \cap (K_1 \cup \ldots \cup K_k))}} \end{aligned}$$

in the latter case. Though the numerators are almost equivalent, and both of the denominators represent computation of a $|(K_{k+1} \cap (K_1 \cup \ldots \cup K_k))|$-dimensional marginal distribution, there is a computational difference between these expressions. While in the first case the denominator represents computation of a marginal from distribution $P_{k+1}$, which is assumed to be low-dimensional, in the latter case one has to marginalize the distribution $(P_1 \triangleleft \ldots \triangleleft P_k)$, whose dimension can be rather high; more precisely, it is $|(K_1 \cup \ldots \cup K_k)|$-dimensional. In practical situations, when the goal is to construct a distribution with dimensionality of several hundreds, these computations become generally intractable (more precisely, no effective algorithms have been found). Therefore, we will concentrate mainly on applications of the operator $\triangleright$. Nevertheless, there are sequences of distributions for which

$$P_1 \triangleright P_2 \triangleright \ldots \triangleright P_n = P_1 \triangleleft P_2 \triangleleft \ldots \triangleleft P_n.$$

holds true. Among such sequences, an important role is played by those that are called perfect (this notion was already introduced in (Jiroušek 1997)). A sequence of probability distributions $P_1, P_2, \ldots, P_n$ is called *perfect* if for all $k = 2, \ldots, n$ the equality

$$P_1 \triangleright \ldots \triangleright P_k = P_1 \triangleleft \ldots \triangleleft P_k,$$

holds true.

It is not difficult to show that the class of Bayesian networks is equivalent to the class of perfect sequences in the following sense:

1. If $P_1, \ldots, P_n$ is perfect then there exists a Bayesian network representing the distribution $P_1 \triangleright \ldots \triangleright P_n$ such that for each variable $X_j$ there exists $k \in \{1, \ldots, n\}$ such that

$$cl(X_j) = (\{X_j\} \cup pa(X_j)) \subset \{X_i\}_{i \in K_k}.$$

2. For each Bayesian network one can construct a perfect sequence $P_1, \ldots, P_n$ such that each $\{X_i\}_{i \in K_k}$ equals some $cl(X_j) = \{X_j\} \cup pa(X_j)$ and $P_1 \triangleright \ldots \triangleright P_n$ equals the distribution represented by the Bayesian network.

In other words, there are simple procedures transforming an arbitrary Bayesian network into a perfect sequence and vice versa; and the distributions defining both structures (i.e., respective conditional distributions defining the Bayesian network and distributions from the generating sequence) are of the same dimensionality. An algorithm for reconstruction of a Bayesian network from a perfect sequence can be found in (Jiroušek et al. 2000). In fact, this algorithm transforms any sequence $P_1, \ldots, P_n$ into a Bayesain network representing the distribution $P_1 \triangleright \ldots \triangleright P_n$. What is more important, from our point of view, is the fact that any Bayesian network can be viewed at as a structure constructed from a perfect sequence of low-dimensional distributions. The exact meaning and importance of this statement can be seen from the following characterization theorem.

**Theorem 2** *A sequence of distributions $P_1, P_2, \ldots, P_n$ is perfect* iff *all the distributions from this sequence are marginals of the distribution $(P_1 \triangleright P_2 \triangleright \ldots \triangleright P_n)$.*

*Proof.* The fact that all distributions $P_k$ from a perfect sequence are marginals of $(P_1 \triangleright P_2 \triangleright \ldots \triangleright P_n)$ was already stated in Theorem 4 in (Jiroušek 1997). It follows from the fact that $(P_1 \triangleright \ldots \triangleright P_k)$ is marginal to $(P_1 \triangleright \ldots \triangleright P_n)$ and $P_k$ is marginal to $(P_1 \triangleleft \ldots \triangleleft P_k)$.

Suppose that for all $k = 1, \ldots, n$, $P_k$ are marginal distributions of $(P_1 \triangleright \ldots \triangleright P_n)$. Then $P_1$ and $P_2$ are consistent, and due to Lemma 1

$$P_1 \triangleright P_2 = P_1 \triangleleft P_2.$$

Since $P_1 \triangleright P_2$ is also marginal to $(P_1 \triangleright \ldots \triangleright P_n)$, it must be consistent with $P_3$, too. Using Lemma 1 again, we get

$$P_1 \triangleright P_2 \triangleright P_3 = P_1 \triangleleft P_2 \triangleleft P_3.$$

However, $P_1 \triangleright P_2 \triangleright P_3$ being marginal to $(P_1 \triangleright \ldots \triangleright P_n)$ must also be consistent with $P_4$ and we can continue in this manner until we achieve that for all $k = 2, \ldots, n$

$$P_1 \triangleright P_2 \triangleright \ldots \triangleright P_k = P_1 \triangleleft P_2 \triangleleft \ldots \triangleleft P_k.$$
□

What is the most important message conveyed by the previous characterization theorem? A distribution defined by a perfect sequence is unique, regardless of which of the two operators ($\triangleleft$ or $\triangleright$) is used. Moreover, considering that low-dimensional distributions $P_k$ are carriers of local information, the constructed multidimensional distribution represents global information, faithfully reflecting all of the local input. The reader can visualize the situation with an analogy to a jigsaw puzzle, whose pieces correspond to individual low-dimensional distributions $P_k$ and whose completed picture corresponds to the distribution $P_1 \triangleright \ldots \triangleright P_n$. In this case, if the picture is properly assembled, each local piece is fully utilized, no piece of information is lost, and no information that is not included in any $P_k$ is added.

There is still another moment worth mentioning to readers who are familiar with the famous *Iterative Proportional Fitting Procedure* (Deming and Stephan 1940, Csiszár 1975). Since the operator $\triangleleft$ describes exactly what is computed by this procedure at each step, $(P_1 \triangleleft \ldots \triangleleft P_n)$ is the distribution computed by the first cycle ($n$ iterative steps) when the procedure starts with the uniform distribution. Moreover, due to the fact that, for perfect sequences, all distributions $P_k$ are marginal to $(P_1 \triangleleft \ldots \triangleleft P_n)$, the iterative process terminates after the $n$-th step. Therefore, for perfect sequences the IPFP terminates after the first cycle.

## 4  MARGINALIZATION

We believe that the apparatus based on composition of distributions from generating sequences is not only an elegant way how to describe multidimensional distributions but we hope it will enable us also to describe necessary computational procedures. These consist mainly from steps performing conditioning and marginalization. Therefore, in this paper we start studying problems connected with marginalization. The goal, however, is not to describe algorithms performing this type of computations (it can be done by any of the famous marginalization procedures proposed for Bayesian networks, see e.g. (Shenoy and Shafer 1990, Shafer and Shenoy 1990)) but to find formulae based on operators of composition describing the resulting marginal distribution.

It is easy to show that generally

$$(P_1 \triangleright P_2)^{(L)} \neq P_1^{(L)} \triangleright P_2^{(L)}.$$

To see it, consider a simple example of composition of two two-dimensional distributions

$$P_1(X_1, X_2) \triangleright P_2(X_2, X_3)$$

that yields, generally, a dependence of variables $X_1$ and $X_3$. Therefore

$$\begin{aligned}(P_1(X_1, X_2) &\triangleright P_2(X_2, X_3))^{(\{1,3\})} \\ &\neq (P_1(X_1, X_2))^{(\{1\})} \triangleright (P_2(X_2, X_3))^{(\{3\})} \\ &= P_1(X_1) P_2(X_3).\end{aligned}$$

Nevertheless, for special situations the following simple assertion (Lemma 2 in (Jiroušek 1997)) presents sufficient conditions under which equality in the above expression holds true.

**Lemma 4** *If $L \supseteq K_1 \cap K_2$ then*

$$(P_1 \triangleright P_2)^{(L)} = P_1^{(L)} \triangleright P_2^{(L)}.$$

□

In the sequel we will primarily concentrate on the simplest case: marginalization of one variable out. From this point of view, the following assertion – an immediate consequence of iterative application of Lemma 1 – is rather interesting:

**Lemma 5** *If $\ell \in K_i$ for some $i \in \{1, 2, \ldots, n\}$ and $\ell \notin K_j$ for all $j \neq i$ then*

$$(P_1 \triangleright P_2 \triangleright \ldots \triangleright P_n)^{[\ell]} = P_1 \triangleright \ldots \triangleright P_{i-1} \triangleright P_i^{[\ell]} \triangleright P_{i+1} \triangleright \ldots \triangleright P_n.$$

□

However, situations in which the variable $\ell$ that is to be eliminated is contained in several distributions, are much more complicated. The solution to this problem is in fact given by the following theorem, which expresses the distribution $P_1 \triangleright \ldots \triangleright P_n$ with the aid of its marginal $(P_1 \triangleright \ldots \triangleright P_n)^{[\ell]}$.

**Theorem 3** *Let $P_1, P_2, \ldots, P_n$ be a generating sequence and $\ell \in K_{i_1} \cap K_{i_2} \cap \ldots \cap K_{i_m}$ for some*

$$\{i_1, i_2, \ldots, i_m\} \subseteq \{1, 2, \ldots, n\}$$

*(assuming $(i_1 < i_2 < \ldots < i_m)$) such that $\ell \notin K_j$ for all $j \in \{1, 2, \ldots, n\} \setminus \{i_1, i_2, \ldots, i_m\}$. Then*

$$P_1 \triangleright P_2 \triangleright \ldots \triangleright P_n = Q_1 \triangleright Q_2 \triangleright \ldots \triangleright Q_n \triangleright Q_{n+1},$$

*where*

$Q_j = P_j$ for all $j \in \{1, \ldots, n\} \setminus \{i_1, \ldots, i_m\}$,
$Q_{i_1} = P_{i_1}^{[\ell]}$,
$Q_{i_2} = (P_{i_1} \otimes_{L_{i_2-1}} P_{i_2})^{[\ell]}$,
$Q_{i_3} = (P_{i_1} \otimes_{L_{i_2-1}} P_{i_2} \otimes_{L_{i_3-1}} P_{i_3})^{[\ell]}$,
$\vdots$
$Q_{i_m} = (P_{i_1} \otimes_{L_{i_2-1}} P_{i_2} \otimes_{L_{i_3-1}} \cdots \otimes_{L_{i_m-1}} P_{i_m})^{[\ell]}$,
$Q_{n+1} = (P_{i_1} \otimes_{L_{i_2-1}} P_{i_2} \otimes_{L_{i_3-1}} \cdots \otimes_{L_{i_m-1}} P_{i_m})$,
and $L_{i_k-1} = (K_1 \cup K_2 \cup \ldots \cup K_{i_k-1}) \setminus \{\ell\}$.

*Proof.* Let us start proving the theorem for $m = 1$. Since $K_1 \cup \ldots \cup K_{i_1-1}$ does not contain $\ell$, we can apply Lemma 3, which yields

$$P_1 \triangleright \ldots \triangleright P_{i_1-1} \triangleright P_{i_1} = (P_1 \triangleright \ldots \triangleright P_{i_1-1}) \triangleright P_{i_1}^{[\ell]} \triangleright P_{i_1}$$
$$= Q_1 \triangleright \ldots \triangleright Q_{i_1} \triangleright P_{i_1}$$

Distribution $(Q_1 \triangleright \ldots \triangleright Q_{i_1})$ is defined for $(X_i)_{i \in (K_1 \cup \ldots \cup K_{i_1}) \setminus \{\ell\}}$ and $(K_1 \cup \ldots \cup K_{i_1}) \setminus \{\ell\}$ contains $K_{i_1} \cap K_j$ for all $j = i_1 + 1, \ldots, n$, because none of these $K_j$ contain $\ell$. Therefore, applying Lemma 2 $(n - i_1)$-times, we get

$$P_1 \triangleright \ldots \triangleright P_{i_1} \triangleright P_{i_1+1} \triangleright \ldots \triangleright P_n$$
$$= Q_1 \triangleright \ldots \triangleright Q_{i_1} \triangleright P_{i_1} \triangleright P_{i_1+1} \triangleright \ldots \triangleright P_n$$
$$= Q_1 \triangleright \ldots \triangleright Q_{i_1} \triangleright P_{i_1+1} \triangleright \ldots \triangleright P_n \triangleright P_{i_1}$$
$$= Q_1 \triangleright \ldots \triangleright Q_{n+1}.$$

Now, assuming the assertion has been proven for $m-1$, let us prove it for $m$. In the following computations we will first use Lemma 3, then Theorem 1, and finally $(n - i_m)$-times Lemma 2.

$$P_1 \triangleright \ldots \triangleright P_{i_m-1} \triangleright P_{i_m} \triangleright \ldots \triangleright P_{i_n}$$
$$= Q_1 \triangleright \ldots \triangleright Q_{i_m-1}$$
$$\triangleright (P_{i_1} \otimes_{L_{i_2-1}} \cdots \otimes_{L_{i_{m-1}-1}} P_{i_{m-1}})$$
$$\triangleright P_{i_m} \triangleright \ldots \triangleright P_{i_n}$$
$$= Q_1 \triangleright \ldots \triangleright Q_{i_m-1}$$
$$\triangleright (P_{i_1} \otimes_{L_{i_2-1}} \cdots \otimes_{L_{i_{m-1}-1}} P_{i_{m-1}})^{[\ell]}$$
$$\triangleright (P_{i_1} \otimes_{L_{i_2-1}} \cdots \otimes_{L_{i_{m-1}-1}} P_{i_{m-1}}) \triangleright P_{i_m}$$
$$\triangleright P_{i_m+1} \triangleright \ldots \triangleright P_{i_n}$$
$$= Q_1 \triangleright \ldots \triangleright Q_{i_m}$$
$$\triangleright (P_{i_1} \otimes_{L_{i_2-1}} \cdots \otimes_{L_{i_{m-1}-1}} P_{i_{m-1}} \otimes_{L_{i_m-1}} P_{i_m})$$
$$\triangleright P_{i_m+1} \triangleright \ldots \triangleright P_{i_n}$$
$$= Q_1 \triangleright \ldots \triangleright Q_{i_m} \triangleright P_{i_m+1} \triangleright \ldots \triangleright P_{i_n}$$
$$\triangleright (P_{i_1} \otimes_{L_{i_2-1}} \cdots \otimes_{L_{i_{m-1}-1}} P_{i_m})$$
$$= Q_1 \triangleright \ldots \triangleright Q_{i_n+1}$$

□

**Theorem 4** *Let $P_1, P_2, \ldots, P_n$ be a generating sequence and $\ell \in K_{i_1} \cap K_{i_2} \cap \ldots \cap K_{i_m}$ for some*

$$\{i_1, i_2, \ldots, i_m\} \subseteq \{1, 2, \ldots, n\}$$

*(assuming $(i_1 < i_2 < \ldots < i_m)$) such that $\ell \notin K_j$ for all $j \in \{1, 2, \ldots, n\} \setminus \{i_1, i_2, \ldots, i_m\}$ then*

$$(P_1 \triangleright P_2 \triangleright \ldots \triangleright P_n)^{[\ell]} = Q_1 \triangleright Q_2 \triangleright \ldots \triangleright Q_n,$$

*where the distributions $Q_1, \ldots, Q_n$ are defined as in Theorem 3.*

*Proof.* In fact, this assertion is a direct consequence of the preceding Theorem, which claims that $Q_1 \triangleright \ldots \triangleright Q_n$ is marginal to $P_1 \triangleright \ldots \triangleright P_n$. The fact that it is the marginal distribution for variables $(X_i)_{i \in (K_1 \cup \ldots \cup K_n) \setminus \{\ell\}}$ immediately follows from the definitions of distributions $Q_k$ since $Q_{n+1}$ is the only distribution in whose domain $\ell$ occurs.

□

We will conclude this section with a simple example, illustrating the marginalization formula of Theorem 4. Consider the following generating sequence:

$$P_1(X_1, X_3), P_2(X_2), P_3(X_1, X_2, X_3, X_4),$$

which is, generally, not perfect. (Nevertheless, it can be perfect if $P_3(X_1, X_2, X_3) = P_1(X_1, X_3)P_2(X_2)$.) The goal of the following computation is to eliminate variable $X_1$.

$$\left(P_1(X_1, X_3) \triangleright P_2(X_2) \triangleright P_3(X_1, X_2, X_3, X_4)\right)^{[1]}$$
$$= P_1(X_3) \triangleright P_2(X_2)$$
$$\triangleright \left(P_1(X_1, X_3) \otimes_{\{1,2,3\}} P_3(X_1, X_2, X_3, X_4)\right)^{[1]}$$
$$= P_1(X_3) P_2(X_2)$$
$$\triangleright \left(P_3(X_2) P_1(X_1, X_3) \triangleright P_3(X_1, X_2, X_3, X_4)\right)^{[1]}$$
$$= P_1(X_3) P_2(X_2)$$
$$\triangleright \left(P_3(X_2) P_1(X_1, X_3) P_3(X_4|X_1, X_2, X_3)\right)^{[1]}$$
$$= P_1(X_3) P_2(X_2)$$
$$\frac{\left(P_3(X_2) P_1(X_1, X_3) P_3(X_4|X_1, X_2, X_3)\right)^{[1]}}{\left(P_3(X_2) P_1(X_1, X_3) P_3(X_4|X_1, X_2, X_3)\right)^{(\{2,3\})}}$$
$$= P_1(X_3) P_2(X_2)$$
$$\frac{P_3(X_2) \left(P_1(X_1, X_3) P_3(X_4|X_1, X_2, X_3)\right)^{[1]}}{P_3(X_2) \left(P_1(X_1, X_3) P_3(X_4|X_1, X_2, X_3)\right)^{(\{2,3\})}}$$
$$= P_1(X_3) P_2(X_2)$$
$$\frac{\left(P_1(X_1, X_3) P_3(X_4|X_1, X_2, X_3)\right)^{[1]}}{\left(P_1(X_1, X_3) P_3(X_4|X_1, X_2, X_3)\right)^{(\{2,3\})}}$$
$$= P_1(X_3) P_2(X_2)$$
$$\frac{\left(P_1(X_1, X_3) P_3(X_4|X_1, X_2, X_3)\right)^{[1]}}{\left(\left(P_1(X_1, X_3) P_3(X_4|X_1, X_2, X_3)\right)^{(\{4\})}\right)^{[1]}}$$
$$= P_1(X_3) P_2(X_2)$$
$$\frac{\left(P_1(X_1, X_3) P_3(X_4|X_1, X_2, X_3)\right)^{[1]}}{\left(P_1(X_1, X_3)\right)^{(\{1\})}}$$
$$= P_2(X_2) \left(P_1(X_1, X_3) P_3(X_4|X_1, X_2, X_3)\right)^{[1]},$$

which was to be expected.

Let us stress once more that this is something quite different from

$$P_1^{[1]} \triangleright P_2 \triangleright \left(P_1 \triangleright P_3\right)^{[1]}$$

which equals

$$P_1(X_3) \triangleright P_2(X_2)$$
$$\triangleright \left(P_1(X_1, X_3) \triangleright P_3(X_1, X_2, X_3, X_4)\right)^{[1]}$$
$$= P_1(X_3) P_2(X_2)$$
$$\triangleright \left(P_1(X_1, X_3) P_3(X_4, X_2|X_1, X_3)\right)^{[1]}$$
$$= P_1(X_3) P_2(X_2)$$
$$\frac{\left(P_1(X_1, X_3) P_3(X_4, X_2|X_1, X_3)\right)^{[1]}}{\left(P_1(X_1, X_3) P_3(X_4, X_2|X_1, X_3)\right)^{(\{2,3\})}}.$$

## 5 CONCLUSIONS

We have presented a contribution to a new apparatus for representations of multidimensional probability distributions, based on composition operators. Although the two basic operators, $\triangleleft$ and $\triangleright$, are defined by almost identical formulae, they substantially differ when used iteratively to constitute multidimensional probabilistic distributions. The difference mostly manifests in the computational complexity of the respective processes.

Within this framework, different generating sequences of low-dimensional probability distributions can be studied (Jiroušek, 1998). In this paper we defined and characterized only the most important class, that of perfect sequences. The main result of this paper, theorem on marginalization for generating sequences, was, however, formulated for general sequences.

Let us conclude the paper with two comments concerning research in the related fields.

We made a rather great deal of effort to characterize sequences that define multidimensional models corresponding to decomposable models. Up to now, we have not received satisfactory results. Naturally, it would be possible to choose perfect sequences $P_1, \ldots, P_n$ for which sets $K_1, \ldots, K_n$ can be ordered to meet the so called *running intersection property* (introduced in (Kellerer, 1964)). This would, however, exclude some situations we want to address. For example, if

$$P_1(X_1, X_2), P_2(X_2, X_3), P_3(X_3, X_4), P_4(X_1, X_4)$$

is perfect, then the distribution

$$P_1(X_1, X_2) \triangleright P_2(X_2, X_3) \triangleright P_3(X_3, X_4) \triangleright P_4(X_1, X_4)$$

is decomposable, because

$$P_1(X_1, X_2) \triangleright P_2(X_2, X_3) \triangleright P_3(X_3, X_4) \triangleright P_4(X_1, X_4)$$
$$= P_1(X_1, X_2) \triangleright P_2(X_2, X_3) \triangleright P_3(X_3, X_4),$$

in spite of the fact that the respective sets $\{1,2\},\{2,3\},\{3,4\},\{1,4\}$ corresponding to the original sequence $P_1, P_2, P_3, P_4$ cannot be ordered to achieve the running intersection property. Another situation we want to address occurs when all the distributions from a generating sequence $P_1, \ldots, P_n$ are uniform. Then the result, the uniform multidimensional distribution, is also decomposable regardless of whether the respective sets $K_1, \ldots, K_n$ can be ordered to meet the running intersection property or not. Thus, the problem of how to specify sequences corresponding to decomposable models is still open.

The second comment goes beyond the probability theory. The operators of composition were also defined for possibilistic distributions (Vejnarová, 1998). Corresponding to the conditioning introduced in (de Cooman, 1997a - 1997c) they are parameterized by a t-norm, nevertheless they manifest a lot of properties which also hold true for probabilistic operators. These properties, for example, make a definition of perfect sequences possible, which can thus be understood as a definition of a possibilistic counterpart of Bayesian networks (Jiroušek et al., 2000). Although a large number of properties are yet to be proven, a chance exists that the composition operators may prove to be tools that will enable us to study multidimensional distributions in both probability and possibility theories, within the framework of a uniform approach.

## Acknowledgments

This work was supported by the grants: Ministry of Education of ČR no. VS96008 and GAČR no. 201/98/1487.